\documentclass[10pt,twocolumn,letterpaper]{article}

\usepackage[pagenumbers]{cvpr} 

\usepackage{tabularx} 

\usepackage{makecell}   
\usepackage{amssymb}    
\usepackage{booktabs}   
\usepackage{amssymb}
\usepackage{bbding}
\usepackage{multirow}

\usepackage[marginal]{footmisc}
\newcommand{\myparagraph}[1]{\vspace{0.1em}\noindent\textbf{#1}}

\usepackage{graphicx}
\usepackage{amsmath}
\usepackage{amssymb}
\usepackage{booktabs}
\usepackage{bm}

\usepackage{cuted}
\usepackage{capt-of}

\usepackage{multirow}
\usepackage{verbatim}
\usepackage{comment}
\usepackage{adjustbox}
\usepackage{wrapfig}

\usepackage{url}            
\usepackage{amsfonts}       
\usepackage{nicefrac}       
\usepackage{microtype}      

\usepackage{array}
\usepackage{xspace}
\usepackage{pifont}
\newcommand{\cmark}{\ding{51}}%
\newcommand{\xmark}{\ding{55}}%

\usepackage[dvipsnames]{xcolor}

\definecolor{cvprblue}{rgb}{0.21,0.49,0.74}
\usepackage[pagebackref,breaklinks,colorlinks,citecolor=cvprblue]{hyperref}

\def\paperID{318} 
\def\confName{CVPR}
\def\confYear{2024}

\title{HOI-M$^3$: Capture Multiple Humans and Objects Interaction within Contextual Environment}

\author {
    Juze Zhang \textsuperscript{\rm 1,\rm 2,\rm *},
    Jingyan Zhang \textsuperscript{\rm 1,\rm *},
    Zining Song \textsuperscript{\rm 1},
    Zhanhe Shi \textsuperscript{\rm 1},
    Chengfeng Zhao \textsuperscript{\rm 1},
    Ye Shi \textsuperscript{\rm 1}, \\
    Jingyi Yu \textsuperscript{\rm 1},
    Lan Xu \textsuperscript{\rm 1},
    Jingya Wang \textsuperscript{\rm 1,\dag}\\
    \textsuperscript{\rm 1} ShanghaiTech University  
    \textsuperscript{\rm 2} University of Chinese Academy of Sciences \\
    \{zhangjz,zhangjy7,songzn,shizhh,zhaochf2022,shiye,yujingyi,xulan1,wangjingya\}@shanghaitech.edu.cn
}

\begin{document}
\maketitle

\footnote{* These authors contributed equally.}
\footnote{\dag Corresponding author.}

\begin{strip}\centering
    \vspace{-40px}
    \captionsetup{type=figure}
    \includegraphics[width=\textwidth]{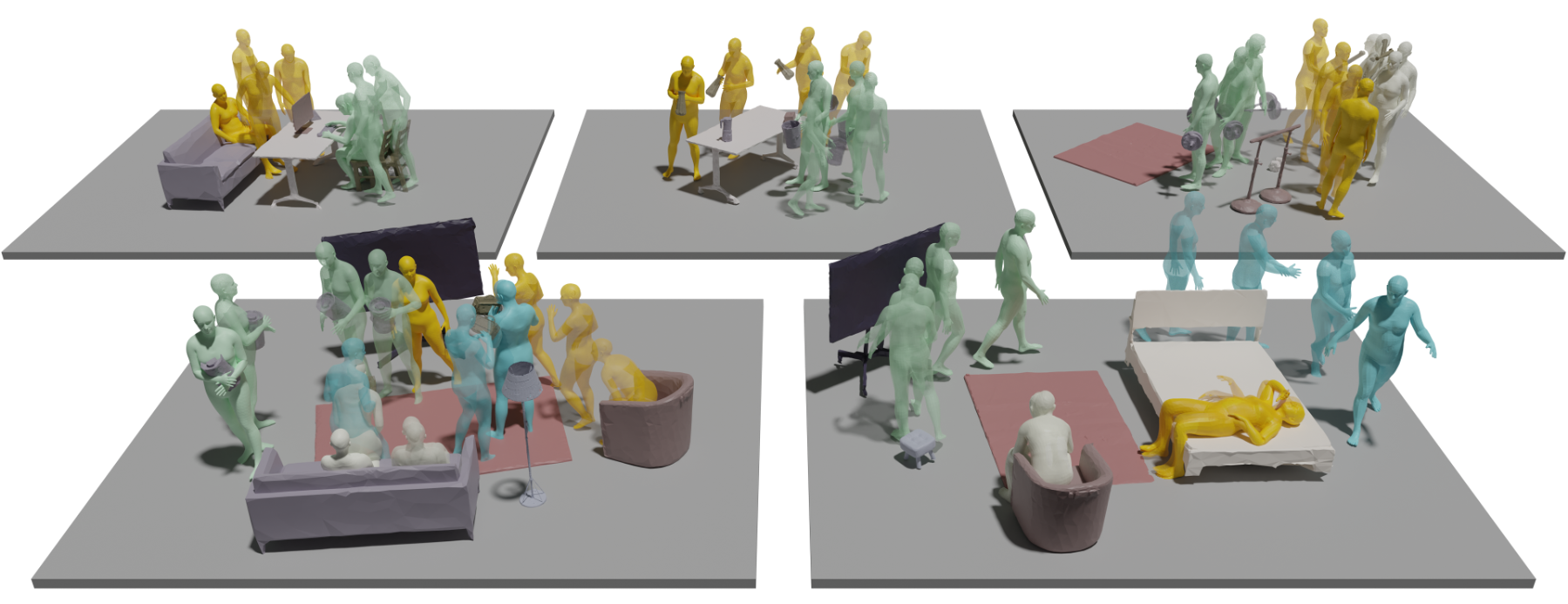}
    \vspace{-20px}
    \captionof{figure}{We meticulously collect a dataset capturing interactions involving multiple humans and multiple objects, named HOI-M$^{3}$. This extensive dataset comprises 181 million video frames recorded from 42 diverse viewpoints, covering a wide range of daily scenarios. It is intended to facilitate various tasks related to human-object interaction perception and generation.}
    \label{fig:teaser}
    \vspace{-5px}
\end{strip}

\begin{abstract}
Humans naturally interact with both others and the surrounding multiple objects, engaging in various social activities. However, recent advances in modeling human-object interactions mostly focus on perceiving isolated individuals and objects, due to fundamental data scarcity. 
In this paper, we introduce HOI-M$^3$, a novel large-scale dataset for modeling the interactions of Multiple huMans and Multiple objects. Notably, it provides accurate 3D tracking for both humans and objects from dense RGB and object-mounted IMU inputs, covering 199 sequences and 181M frames of diverse humans and objects under rich activities. With the unique HOI-M$^3$ dataset, we introduce two novel data-driven tasks with companion strong baselines: monocular capture and unstructured generation of multiple human-object interactions.
Extensive experiments demonstrate that our dataset is challenging and worthy of further research about multiple human-object interactions and behavior analysis. Our HOI-M$^3$ dataset, corresponding codes, and pre-trained models will be disseminated to the community for future research, which can be found at \url{https://juzezhang.github.io/HOIM3_ProjectPage/}

\end{abstract}

\section{Introduction}
\label{sec:intro}

Modeling human behaviors with surrounding objects within contextual environments is a fundamental task in the vision community,  enabling numerous applications for gaming, embodied AI, robotics, and VR/AR. Capturing such human-object interactions recently received substantive attention.

With the aid of a wide range of available datasets~\cite{ionescu2013human3,mahmood2019amass}, these years have witnessed the huge progress of data-driven human motion modeling, from motion capture (MoCap)~\cite{VIBE_CVPR2020,zhang20204d,zhang2023ikol,kolotouros2019spin,li2021hybrik,ren2023lidar} to recently emerging motion generation (MoGen)~\cite{tevet2022human,dabral2022mofusion, karunratanakul2023gmd, zhang2020generating, zhang2020place, zhang2021we,li2023controllable,wang2023physhoi,diller2023cg,zhang2024force,peng2023hoi,cui2024anyskill,jiang2024scaling,araujo2023circle,pi2023hierarchical,dai2024interfusion,jiang2021hand,yi2023mime}. Yet, the further 3D modeling of human-object interactions (HOI) significantly falls behind, mainly due to the scarcity of data.
Specifically, recent available MoCap datasets~\cite{bhatnagar2022behave,huang2022intercap,zhang2023neuraldome} for HOI mostly focus on interactions between a single human and individual objects. Hence the data-driven MoCap advances~\cite{zhang2020phosa,xie2022chore,xie2023visibility,ijcai2023p100} for HOI are restricted to the single-person scenarios. They fall short of modeling the interactions between multiple humans and objects, which is crucial for a comprehensive understanding of how we humans and objects interact in social settings.

\begin{figure*}[t!]
  \centering
  \includegraphics[width=0.96\linewidth]{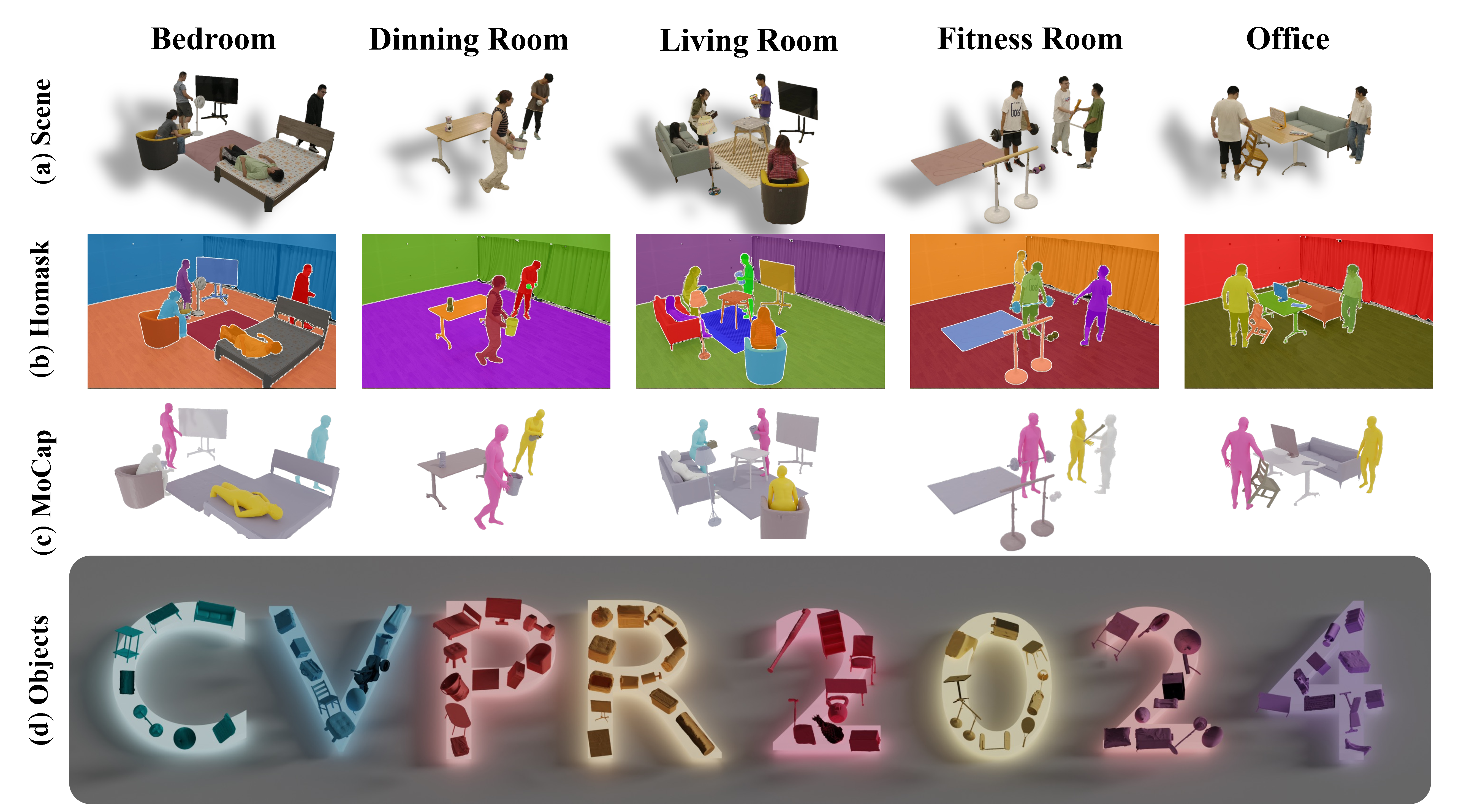}
    \captionof{figure}{\textbf{Overview of HOI-M$^3$}. (a) HOI-M$^3$ across five daily scenarios(Bedroom, Dinning Room, Living Room, Fitness Room, Office), (b) annotated masks corresponding to each subject(human, object), (c) tracking of multiple humans and multiple objects, (d) significant number of pre-scanned object meshes.}
  \label{fig:Galley}
\end{figure*}

However, accurately capturing the motions of multiple humans and objects remains challenging due to the severe occlusion, especially for daily interactions within contextual environments. First, it usually requires dome-like dense cameras~\cite{collet2015high,zhang2023neuraldome} and even object-mounted Inertial Measurement Units (IMUs)~\cite{XSENS} to provide sufficient motion observations. Second, even based on such dense and multi-modal input, an accurate capture method remains far-reaching. It requires a series of tedious and time-consuming stages, ranging from pre-processing, i.e., human-object segmentation and sensor alignment, to a robust joint optimization process, or even manual correction for those extremely occluded cases. These challenges hinder existing HOI methods to explore the multi-human and multi-object scenarios, and hence solving this data scarcity is a long-standing and urgent issue.  


To tackle these challenges, in this paper, we present \textit{HOI-M$^3$} -- a novel and timely dataset for modeling the interactions of \textbf{M}ultiple hu\textbf{M}ans and \textbf{M}ultiple objects, as illustrated in Figure~\ref{fig:teaser}. We adopt a dense and hybrid capture setting with a robust human-object capture pipeline to accurately track the 3D motions of various humans and objects, providing more than 199 human-object inter-acting sequences covering 90 diverse 3D objects and 31 human subjects (20 males and 11 females) across various environment. Noteworthy features of our HOI-M$^3$ dataset include 1) \textbf{Multiple Humans and Objects}: Each sequence involves a minimum number of 2 persons and 5 objects, which, to the best of our knowledge, is the first real-world 3D multiple human-object datasets with accurate 3D MoCap. 2) \textbf{High Quality}: Sequences are recorded within daily-style rooms with 42 synchronized camera views, and inertial measurement units (IMUs) are embedded in each pre-scanned object to ensure accurate human-object tracking labels. 3) \textbf{Large Size and Rich Modality}: Our dataset records over 20 hours of interactions with both RGB and inertial sensors, providing segmentation annotations, pre-scanned object geometry, and accurate HOI tracking labels.

Note that our HOI-M$^3$ dataset is the first of its kind to open up the research direction for data-driven multiple human-object motion capture or even synthesis. The rich annotations and multi-modality of our dataset also bring huge potential for future direction for HOI modeling and behavior analysis. 
To this end, based on our novel HOI-M$^3$ dataset, we provide two strong baseline methods for two novel downstream tasks: 1) monocular capture of multiple HOI; 2) unstructured generation of multiple HOI. For the former, we introduce a novel single-shot learning-based method to estimate multi-person and multi-object 3D poses. For the latter, we tailor the diffusion models~\cite{ho2020denoising,li2023ego} into the realm of generating intricate social interactions. 
We conduct detailed evaluations of our dataset and companion baseline methods and provide preliminary results to indicate that capturing or generating vivid motions of multiple human-object interactions remains be challenging a direction. Our HOI-M$^3$ dataset consistently serves as a data foundation and reliable benchmark to facilitate future exploration.
To summarize, our main contributions include:
\begin{itemize} 
	\setlength\itemsep{0em}
 
	\item We contribute a comprehensive motion dataset for multi-person and multi-object interactions (HOI-M$^3$), featuring high quality, large size, and rich modality.
 
	\item We adopt a robust joint optimization to accurately track the 3D motions of both the humans and objects in our dataset, from dense RGB and object-mounted IMU inputs.
 
 	\item We introduce two novel tasks with companion baselines: monocular multiple HOI capturing and generation, showcasing their potential for further exploration.

        \item We will release our dataset, our code and pre-trained models to stimulate the research of human-object interactions.

\end{itemize} 


\section{Related Works}

\noindent\textbf{Single Human and Object Interaction.}
Several recent studies\cite{taheri2020grab,zhang2020phosa,bhatnagar2022behave,huang2022intercap,xie2022chore,zhang2023neuraldome,xie2023visibility,sun2021neural,zhao2023m} have tackled the vital challenge of integrated modeling for interactions involving the entire human body. 
Recently, a plethora of works have delved into the examination of this relationship, employing a range of interaction constraints such as spatial arrangements\cite{zhang2020phosa}, contact maps\cite{hassan2019resolving,taheri2020grab,bhatnagar2022behave,huang2022intercap，yan2023cimi4d}, occlusion\cite{xie2022chore,Xie_2023_CVPR}, and adherence to physical plausibility\cite{Yi_2022_CVPR}.
The most relevant works\cite{weng2021holistic} aim to jointly estimate human pose and scene geometry from a single RGB image.
However, this approach only considers the spatial layout between a single person and multiple objects, without taking into account movable objects.
Nevertheless, the interactions we engage in daily are intricate and diverse.
Current methods attempt to model these interactions by focusing on single  interaction, resulting in a biased representation.
Comparably, we propose a novel paradigm modeling the interactions between multiple human and object interactions. 


\noindent\textbf{Human Interaction with Static Scene.}
Another kind of work considers the holistic scene for interactions. Unlike studies focusing on body-object interactions, these works typically represent the entire environment as a static CAD model, concentrating solely on interactions involving a singular human.
Pioneer works such as PiGraphs\cite{savva2016pigraphs}, captured with RGB-D sensors, suffer from inaccurately reconstructed scenes. Succeed work, PROX\cite{hassan2019resolving}, reconstruct human motions within scene from monocular RGB-D but still exhibit noticeable inferiority. GTA-IM~\cite{cao2020long} exploits the game engine to collect a synthetic dataset with restricted HSIs and scene diversities. Recent work HUMANISE~\cite{wang2022humanise} synthesizes extensive HSIs by aligning high-quality motions with real-world 3D scenes in ScanNet~\cite{dai2017scannet}. In conclusion, these methods consider human activities within static surroundings, overlooking broader engagements with dynamic objects.

\noindent\textbf{Interaction Datasets.}
Numerous datasets are available for the isolated study of humans~\cite{mehta2018single, ionescu2013human3,vonMarcard2018} but few address the contextual environment in which humans operate. A limited number of recent works~\cite{savva2016pigraphs, taheri2020grab, hassan2019resolving, bhatnagar2022behave, huang2022intercap, dabral2021gravicap, xu2021d3dhoi, zhang2022couch, zhang2023neuraldome, jiang2023fullbody, hassan2021stochastic, Huang:CVPR:2022, cao2020long, wang2022humanise, araujo2023circle, li2023object} have focused on capturing humans with surrounding objects and scenes. Recent datasets capture HOI using various methods such as optical markers~\cite{taheri2020grab, li2023object, zhang2023neuraldome}, sparse RGB sensors~\cite{bhatnagar2022behave, huang2022intercap}, IMUs~\cite{zhang2022couch, jiang2023fullbody}, and even 76 RGB sensors~\cite{zhang2023neuraldome}, yet still fall short in addressing the complexities of real-world scenarios. Datasets focusing on HSI capture interactions within static scenes~\cite{hassan2021stochastic, wang2022humanise} using RGBD~\cite{hassan2019resolving} or synthesizing with a Meta Quest 2 headset~\cite{araujo2023circle} to construct scene constraints for interactions. Consequently, the existing literature on interactions involving multiple humans and objects is notably scarce. To bridge this gap, we propose HOI-M$^3$ for capturing multiple human and object interactions within a contextual environment, facilitating various perception or generative HOI tasks.

\section{HOI-M$^3$ Dataset}

\begin{table*}[t!]\small
    \centering
    \setlength{\tabcolsep}{1.4pt}
        \begin{tabular}{llccccccccc}
            \toprule
            \makecell{ } & Datasets & \makecell{multi-person}                          & \makecell{multi-object} & \makecell{dynamic object} & \makecell{\# Recording} & \makecell{\# Frame(M)} & \makecell{Resolution} & \makecell{Fps} & \makecell{ Obj. Num.}      &\makecell{Social interact}\\
            \midrule
            \multirow{10}{*}{\makecell{HOI} }        & PiGr~\cite{savva2016pigraphs}         & \makecell{\textcolor{red}{\xmark}}   & \makecell{\textcolor{green}{\cmark}} & \makecell{\textcolor{red}{\xmark}}   & \makecell{2h}          & \makecell{0.1}        & \makecell{960 × 540}         & \makecell{15}  & \makecell{NA} &\makecell{\textcolor{red}{\xmark}}\\
                                                     & GRAB~\cite{taheri2020grab}           & \makecell{\textcolor{red}{\xmark}}   & \makecell{\textcolor{red}{\xmark}}   & \makecell{\textcolor{green}{\cmark}} & \makecell{3.75~h}      & \makecell{1.62}         & \makecell{NA}                & \makecell{120}  & \makecell{51}
            &\makecell{\textcolor{red}{\xmark}}\\
                                                     & BEHAVE~\cite{bhatnagar2022behave}     & \makecell{\textcolor{red}{\xmark}
            }                                        & \makecell{\textcolor{red}{\xmark}}                              & \makecell{\textcolor{green}{\cmark}} & \makecell{2~h}        & \makecell{0.15}       & \makecell{2048 × 1536} & \makecell{30}         & \makecell{20}                                                 &\makecell{\textcolor{red}{\xmark}}\\
                                                     & InterCap~\cite{huang2022intercap}    & \makecell{\textcolor{red}{\xmark}}   & \makecell{\textcolor{red}{\xmark}}   & \makecell{\textcolor{green}{\cmark}} & \makecell{6~h}         & \makecell{0.07}       & \makecell{1920 × 1080}       & \makecell{30}  & \makecell{10}
            &\makecell{\textcolor{red}{\xmark}}\\
                                                     & GraviCap~\cite{dabral2021gravicap}    & \makecell{\textcolor{red}{\xmark}}   & \makecell{\textcolor{green}{\cmark}} & \makecell{\textcolor{green}{\cmark}} & \makecell{NA}           & \makecell{0.005}      & \makecell{1200 × 877}   & \makecell{24}   & \makecell{4}  &\makecell{\textcolor{red}{\xmark}}\\
                                                     & D3D-HOI~\cite{xu2021d3dhoi}           & \makecell{\textcolor{red}{\xmark}}   & \makecell{\textcolor{red}{\xmark}}   & \makecell{\textcolor{green}{\cmark}} & \makecell{0.58}        & \makecell{0.006}      & \makecell{1280 × 720}   & \makecell{3}   & \makecell{8}  &\makecell{\textcolor{red}{\xmark}}\\
                                                     & COUCH~\cite{zhang2022couch}          & \makecell{\textcolor{red}{\xmark}}   & \makecell{\textcolor{red}{\xmark}}   & \makecell{\textcolor{red}{\xmark}}   & \makecell{3~h}         & \makecell{0.324}      & \makecell{2048 × 1536}                 & \makecell{30}   & \makecell{4}  &\makecell{\textcolor{red}{\xmark}}\\
                                                     & NeuralDome~\cite{zhang2023neuraldome} & \makecell{\textcolor{red}{\xmark}}   & \makecell{\textcolor{red}{\xmark}}   & \makecell{\textcolor{green}{\cmark}} & \makecell{4.3~h}       & \makecell{71}         & \makecell{3840 × 2160}        & \makecell{60}  & \makecell{23} &\makecell{\textcolor{red}{\xmark}}\\
                                                     & CHAIRS~\cite{jiang2023fullbody}       & \makecell{\textcolor{red}{\xmark}}   & \makecell{\textcolor{red}{\xmark}}   & \makecell{\textcolor{green}{\cmark}} & \makecell{17.3h}       & \makecell{1.86}       & \makecell{960 × 540}    & \makecell{30}  & \makecell{81} &\makecell{\textcolor{red}{\xmark}}\\
                                                     & OMOMO~\cite{li2023object}             & \makecell{\textcolor{red}{\xmark}}   & \makecell{\textcolor{red}{\xmark}}   & \makecell{\textcolor{green}{\cmark}} & \makecell{10h}         & \makecell{NA}          & \makecell{NA}       & \makecell{NA}   & \makecell{15} &\makecell{\textcolor{red}{\xmark}}\\
            \midrule
            \multirow{6}{*}{\makecell{HSI}}          & PROX~\cite{hassan2019resolving}       & \makecell{\textcolor{red}{\xmark}}   & \makecell{\textcolor{green}{\cmark}} & \makecell{\textcolor{red}{\xmark}}   & \makecell{NA}           & \makecell{0.1}        & \makecell{1920 × 1080}       & \makecell{30}  & \makecell{NA} &\makecell{\textcolor{red}{\xmark}}\\
                                                     & SAMP~\cite{hassan2021stochastic}     & \makecell{\textcolor{red}{\xmark}}   & \makecell{\textcolor{green}{\cmark}} & \makecell{\textcolor{red}{\xmark}}   & \makecell{~100min}     & \makecell{0.185}      & \makecell{NA}       & \makecell{30}  & \makecell{7}  &\makecell{\textcolor{red}{\xmark}}\\
                                                     & RICH~\cite{Huang:CVPR:2022}           & \makecell{\textcolor{red}{\xmark}} & \makecell{\textcolor{red}{\xmark}}         & \makecell{\textcolor{red}{\xmark}}         & \makecell{NA}           & \makecell{0.577}      & \makecell{4096 × 2160} & \makecell{30}  & \makecell{NA} &\makecell{\textcolor{red}{\xmark}}\\
                                                     & GTA-IM~\cite{cao2020long}             & \makecell{\textcolor{green}{\cmark}} & \makecell{\textcolor{red}{\xmark}}   & \makecell{\textcolor{red}{\xmark}}   & \makecell{NA}           & \makecell{1}          & \makecell{1920 × 1080} & \makecell{NA}   & \makecell{NA}  &\makecell{\textcolor{red}{\xmark}}\\
                                                     & HUMANISE~\cite{wang2022humanise}      & \makecell{\textcolor{red}{\xmark}}   & \makecell{\textcolor{red}{\xmark}}   & \makecell{\textcolor{red}{\xmark}}   & \makecell{11.11h}      & \makecell{1.2}        & \makecell{512 × 512}    & \makecell{NA}  & \makecell{NA} &\makecell{\textcolor{red}{\xmark}}\\
                                                     & CIRCLE~\cite{araujo2023circle}        & \makecell{\textcolor{red}{\xmark}}   & \makecell{\textcolor{red}{\xmark}}   & \makecell{\textcolor{red}{\xmark}}   & \makecell{10h}         & \makecell{4.31}       & \makecell{NA}                & \makecell{120} & \makecell{NA} &\makecell{\textcolor{red}{\xmark}}\\
            \midrule
            & Ours     & \makecell{\textcolor{green}{\cmark}}                            & \makecell{\textcolor{green}{\cmark}} & \makecell{\textcolor{green}{\cmark}} & \makecell{\textbf{20~h}}       & \makecell{\textbf{180.5}}         & \makecell{3840× 2160} & \makecell{60}                & \makecell{\textbf{90}}                  &\makecell{\textcolor{green}{\cmark}}\\
            \bottomrule
        \end{tabular}
    \caption{\textbf{Dataset Comparisons}. We compare our proposed HOI-M$^3$ dataset with existing publicly available HOI/HSI datasets. HOI-M$^3$ exhibits the largest scale of interactions in terms of the number of frames (\#Frame) and recording time. It is the first dataset featuring multi-person and multi-object tracking. "Obj. Num." represents the number of objects.}
    \label{Dataset Comparion}
\end{table*}

\subsection{Overview}



We present the HOI-M$^3$ dataset, designed to capture multiple human-object interactions within a contextual environment. As depicted in Table~\ref{Dataset Comparion}, the HOI-M$^3$ dataset encompasses interactions with large size and rich modality involving multiple humans and objects as shown in Figure~\ref{fig:Galley}. It includes 181 million frames featuring 46 subjects engaged in interactions with 90 objects. The dataset provides dense-view coverage at a resolution of 4K and a frame rate of 60 Fps. We highlight the dataset's advantages in terms of recording times, sequence frames, object count, and interaction types, addressing gaps in previous interaction datasets.

\subsection{Data Capture System}
To assemble the HOI-M$^3$ dataset, we deployed 42 Z CAM cinema cameras. Additionally, inertial measurement units (IMUs) were strategically embedded into each pre-scanned object to ensure precision in human-object tracking tasks. Subsequently, publicly accessible tools~\cite{realitycapture} were employed for the estimation of intrinsic camera parameters and extrinsic camera parameters.
\label{sec:formatting}


\subsection{Dataset Process Pipeline}

\textbf{Data Annotation.} 
To collect an extensive and diverse dataset, we conducted pre-scans of 90 commonly used everyday objects spanning various categories. Polycam \cite{Polycam} was employed as our scanning tool for this purpose. We applied segmentation to both humans and objects within the scenes, utilizing the recent Segment Anything Model (SAM)~\cite{kirillov2023segment}. Leveraging SAM's capabilities, we collaborated with professional human annotators to annotate the initial frame of each camera view, ensuring thorough segmentation and broadcasting the entire sequence. Our dataset will be accessible for research purposes.

\noindent \textbf{Synchronization and Calibration.} 
To achieve synchronization between RGB frames and the IMU signal, we instruct the subject to perform a controlled jump at the start of each capture sequence. Subsequently, we manually identify the peaks in both the IMU signal and RGB frames, ensuring temporal alignment between the visual and inertial information. To calibrate the rigid offset between the IMU and RGB systems, we follow these steps: Initially, an IMU is embedded within a typical pre-scanned object, and a human annotator marks three corresponding points in each camera view to determine the object's pose using a triangulation algorithm. This process provides an estimate of the IMU-to-RGB rigid rotation offset, facilitating the extraction of per-frame rotations from IMU signals.
\subsection{Human Motion Capture}

\noindent\textbf{Detection and Matching}. With synchronized and calibrated multi-view videos, we utilize the off-the-shelf 2D pose detection model ViTPose~\cite{xu2022vitpose} to identify 2D human keypoints. Subsequently, we perform a matching process to establish cross-view correspondences for humans observed from different views. Specifically, we formulate a cross-view affinity matrix and address the multi-view matching problem using an established algorithm~\cite{dong2021fast}. Following the matching process, the 3D keypoint trajectories for each entity can be reconstructed through triangulation.

\noindent\textbf{SMPL Fitting}.
We employed SMPL \cite{SMPL:2015} as the underlying body model, offering a differentiable function $\mathcal{M}(\cdot)$ to manipulate a mesh created by artists, consisting of $N=6090$ vertices and $K=24$ joints. Note that we utilized the off-the-shelf toolbox Easymocap~\cite{easymocap} for fitting a parametric model to 3D keypoint.

\begin{figure*}[t!]
  \centering
  \includegraphics[width=0.95\linewidth]{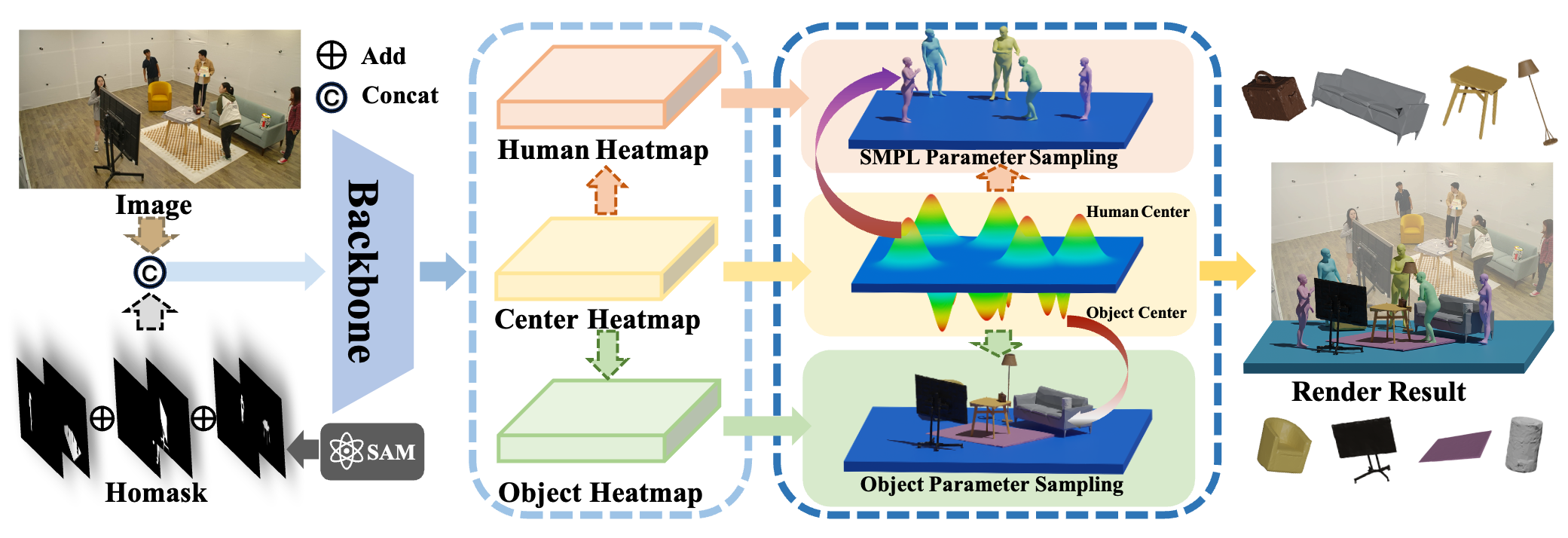}
 \caption{\textbf{Monocular One-Stage Multiple HOI Capturing Pipeline.} Given an input image, the pipeline predicts multiple maps: 1) the human-object center heatmap predicts the probability of the human's root position or object's center position, 2) the human mesh map contains the SMPL parameters and root depth, 3) the object mesh map contains the object 6D pose parameters and center depth. Through the sampling process, multiple humans and objects can be captured within a single forward process.}
  \label{fig:pipeline}
\end{figure*}

\subsection{Inertial-aid Multi-object Tracking}
With the aim of developing a cost-effective scheme that facilitates accurate tracking, we propose an inertial-aided multi-object tracking method. In the context of 3D space, each object is uniquely characterized by its 3D translation $\mathbf{T}\in \mathbb{R}^3$ and 3D rotation $\mathbf{R}\in \mathcal{SO}(3)$. For a rigid object mounted with an IMU, we can easily obtain each frame of rotation. However, the drift error of IMUs tends to reduce confidence as the duration of use extends. Additionally, calibration errors further exacerbate the decline in precision during object tracking. To achieve precise object tracking, we employ an optimization scheme that effectively estimates the object's rotation and translation. We assume the IMU provides plausible rotation $R^{\text{IMU}}_t$, thus we only need to optimize the translation $T_t$ and rotation offset $R^{\text{off}}_t$. The 3D location of the object mesh on a per-frame basis is represented as,
\begin{equation}
V_t^j( R^{\text{IMU}}_t, R^{\text{off}}_t,T_t) = R^{\text{off}}_t R^{\text{IMU}}_t \mathcal{O}(c_j) + T_t,
\end{equation}
where $\mathcal{O}(c_j)$ represents the category $c_j$ mesh template. $T_t$ and $R_t$ are the rigid translation and rotation with respect to its pre-scanned template on each frame $t$. $R^{\text{off}}_t$ is used to eliminate the calibration offset. We use the following four constraints: the object's mask constraint $E_{\text{mask}}$ and offscreen loss $E_{\text{offscreen}}$:
\begin{equation} \label{eq:tracking}
\begin{aligned}
R^{\text{off}}_t, T_t =  \underset{R,T}{{\arg\min} } (& \lambda_{\text{mask}} E_{\text{mask}} + \lambda_{\text{offscreen}} E_{\text{offscreen}} \\
&+ \lambda_{\text{collision}} E_{\text{collision}} +   \lambda_{\text{smt}} E_{\text{smt}}  ),
\end{aligned}
\end{equation}
where $\lambda_{\text{mask}}$, $\lambda_{\text{offscreen}}$, $\lambda_{\text{collision}}$ and $ \lambda_{\text{smt}}$ are coefficients of energy terms.

\noindent \textbf{Human object mask.} Due to the lack of powerful object keypoint detection tools, human and object masks are the strongest evidence for object tracking. Thus, we impose the mask loss as follows:
\begin{equation}
\begin{aligned}
E_{\text{homask}}=  \| \sum_{v=1}^{42} ( I_v^{\text{homask}} - DR(\mathcal{O}(c_j) , R^{\text{IMU}}_t, T_t )  \|_2^2 ,
\end{aligned}
\end{equation}
where $ DR $ denotes differentiable rendering~\cite{kato2018neural}, $I_j^{\text{hmask}}$ and $I_v^{\text{omask}}$ denote human and object masks of $v$-th view computed from the SAM model.

\noindent \textbf{Offscreen loss.} To prevent the degenerate solution of moving the object offscreen, we regularize object within the field of all camera views as:
\begin{equation}
\begin{aligned}
E_{\text{offscreen}} &= \sum_{v=1}^{42}\sum_{[x_v,y_v,z]}\\
& \big[\max(x_v-1, 0) + \max(-1-x_v, 0)\\
& + \max(y_v-1, 0) + \max(-1-y_v, 0)\\
& + \max(-z_v, 0) + \max(z_v-Z_{far}, 0)\big],
\end{aligned}
\end{equation}
where $x_v,y_v$ represents the projected object mesh $DR(V_t)$ in the $v$-th view image coordinate normalized to $[-1,1]$, $z$ is the estimated depth of object and $Z_{far}=200$ is a hyperparameter of the far plane.

\noindent \textbf{Collision constraint.}
Encouraging close proximity between individuals and objects can exacerbate the issue of instances occupying the same 3D space. To tackle this challenge, we introduce a penalty for poses that result in human and/or object interpenetration, employing the collision loss, as introduced in~\cite{zen2012exploiting, tzionas2016capturing}.

\noindent \textbf{Smooth constraint.} Per-frame fitting will damage the smoothness of IMU signal. To encourage the motion estimated rotation to be as smooth as the original IMU signal, we introduce a smooth constraint, which can be written as follows:
\begin{equation}
\begin{aligned}
E_{\text{smt}} = \max(0, & \|  (R^{\text{off}}_t R^{\text{IMU}}_t )^{-1} R^{\text{off}}_{t+1} R^{\text{IMU}}_{t+1} \|_2   \\
  & -  \|  (R^{\text{IMU}}_t )^{-1} R^{\text{IMU}}_{t+1} \|_2 ).
\end{aligned}
\end{equation}


\section{Downstream Tasks}

Leveraging our dataset, we meticulously devised two robust baseline methods for two novel downstream tasks: monocular capture of multiple HOI (Section~\ref{section:Capture}) and unstructured generation of multiple HOI(Section~\ref{section:Generation}).

\subsection{Monocular Multiple HOI Capture}\label{section:Capture}

Monocular perception stands as one of the foundational tasks in visual understanding. In this section, we elucidate how HOI-M$^3$ enhances the robustness analysis for scenarios involving multiple humans and multiple objects. To this end, we propose a one-stage method designed to estimate multi-person and multi-object 3D poses in general scenes from monocular inputs, as illustrated in Figure~\ref{fig:pipeline}. Given an image $I$, our pipeline reconstructs the body meshes of all individual persons and the 6D poses of all objects within $I$. We depict each person or object instance as a singular point in image coordinates. 
With this representation, the pipeline predicts multiple maps.

\noindent \textbf{Human object center heatmap.} 
We used a heatmap representing the 2D human body center and object center in the image. Here we denote the root joint as body center points and object center of mask as the center points. Each center is represented as a Gaussian distribution in the human object center heatmap. 

\noindent \textbf{Human mesh map.}
Following prior works \cite{zhang2021body}, we utilize the body mesh map to reconstruct the body mesh. Specifically, upon detecting a positive response in the root heatmap, we perform regression on the body mesh representation using features from the corresponding feature position, as illustrated in Figure~\ref{fig:pipeline}. For human depth, we employ perspective camera models to project the absolute camera-centric depth of each person \cite{zhang2022mutual}. Consequently, we regress the root depth similar to the body parameters. Adopting a method from a previous study~\cite{zhen2020smap}, we normalize the root depth by the size of the field of view (FoV) as follows:

\begin{equation} 
\begin{aligned}
\hat{Z} = Z \frac{w}{f},
\end{aligned}
\end{equation}
where $\hat{Z}$ is the normalized depth, $Z$ is the original depth, $f$ is the focal length, and $w$ is the image width in pixels.

\noindent \textbf{Object mesh map.} Different from previous multi-stage methodologies, we incorporate object information into a feature map that utilizes the object mesh map for the reconstruction of the object's 6D pose, represented by $R \in \mathbb{R}^{3\times3}$ and $T \in \mathbb{R}^{3}$. To enhance training stability, we employ a 6D rotation representation for the rotation parameters. Analogous to the human branch, we also devise an object depth map to predict absolute depths for all objects in the image, as illustrated in Figure~\ref{fig:pipeline}.

\noindent \textbf{Loss Functions.} 
To supervise the network, we employ individual loss functions for different maps. The network is ultimately supervised by the weighted sum of several loss functions, formulated as follows:
\begin{equation} \label{eq:depth}
\begin{aligned}
L_{\text{sum}} & = \lambda_{\text{theta}} L_{\text{theta}} +  \lambda_{\text{beta}} L_{\text{beta}} + \lambda_{\text{object}} L_{\text{object}} + \\
 & \lambda_{\text{3D}} L_{\text{3D}} + \lambda_{\text{2D}} L_{\text{2D}} + \lambda_{\text{hm}} L_{\text{hm}} + \lambda_{\text{depth}} L_{\text{depth}},
\end{aligned}
\end{equation}
where $ L_{\text{theta}}$, $L_{\text{beta}}$, $L_{\text{object}}$ represent the $\ell_1$ norm between the predicted and ground truth SMPL parameters as well as the object, respectively. $L_{\text{2D}}$ is the 2D keypoints loss that minimizes the distance between the 2D projection from 3D keypoints and ground truth 2D keypoints. $L_{\text{hm}}$ is the mean squared error (MSE) of the predicted and ground truth 2D center keypoint computed from the projected 2D keypoints. Lastly, $\lambda(\cdot)$ denotes the corresponding loss weights. Due to page limitation, we have to defer more details of the loss terms in the Appendix.



\subsection{Multiple Interaction Generation}\label{section:Generation}
HOI-M$^{3}$ offers a wealth of diverse interaction sequences with synchronized ground truth capture. Motivated by the recent remarkable progress in MoGen tasks, we illustrate how our dataset contributes to this field. Currently, generative models have mainly been employed to generate single-person motion diffusion or motion for single objects, with no existing model for the generation of motions involving multiple people and objects. We have meticulously designed a diffusion model for the generation of motions involving multiple people and objects to address this gap.

\noindent \textbf{Multiple HOI representation.}
The parameters for individuals and objects are denoted as $ x = [ x_1, x_2, ... ,x_N]$, where $x_i \in \mathbb{R}^{88}$ encompasses human pose $\theta_i \in \mathbb{R}^{24 \times 3}$, human shape $\beta_i \in \mathbb{R}^{10}$, human global translation $T^h_i \in \mathbb{R}^{3}$, human global orientation $R^h_i \in \mathbb{R}^{3}$ by axis-angle representation, object translation $T^o_i \in \mathbb{R}^{3}$, and object pose $R^o_i \in \mathbb{R}^{3}$. Given that the maximum number of individuals in the HOI dataset does not exceed 5, and the number of objects does not exceed 10, the dimension of our diffusion model is $\mathbb{R}^{500}$, with the first 440 parameters representing 5 people and the last 60 parameters representing 10 objects.

\begin{figure}[t]
    \centering
    \includegraphics[width=\linewidth]{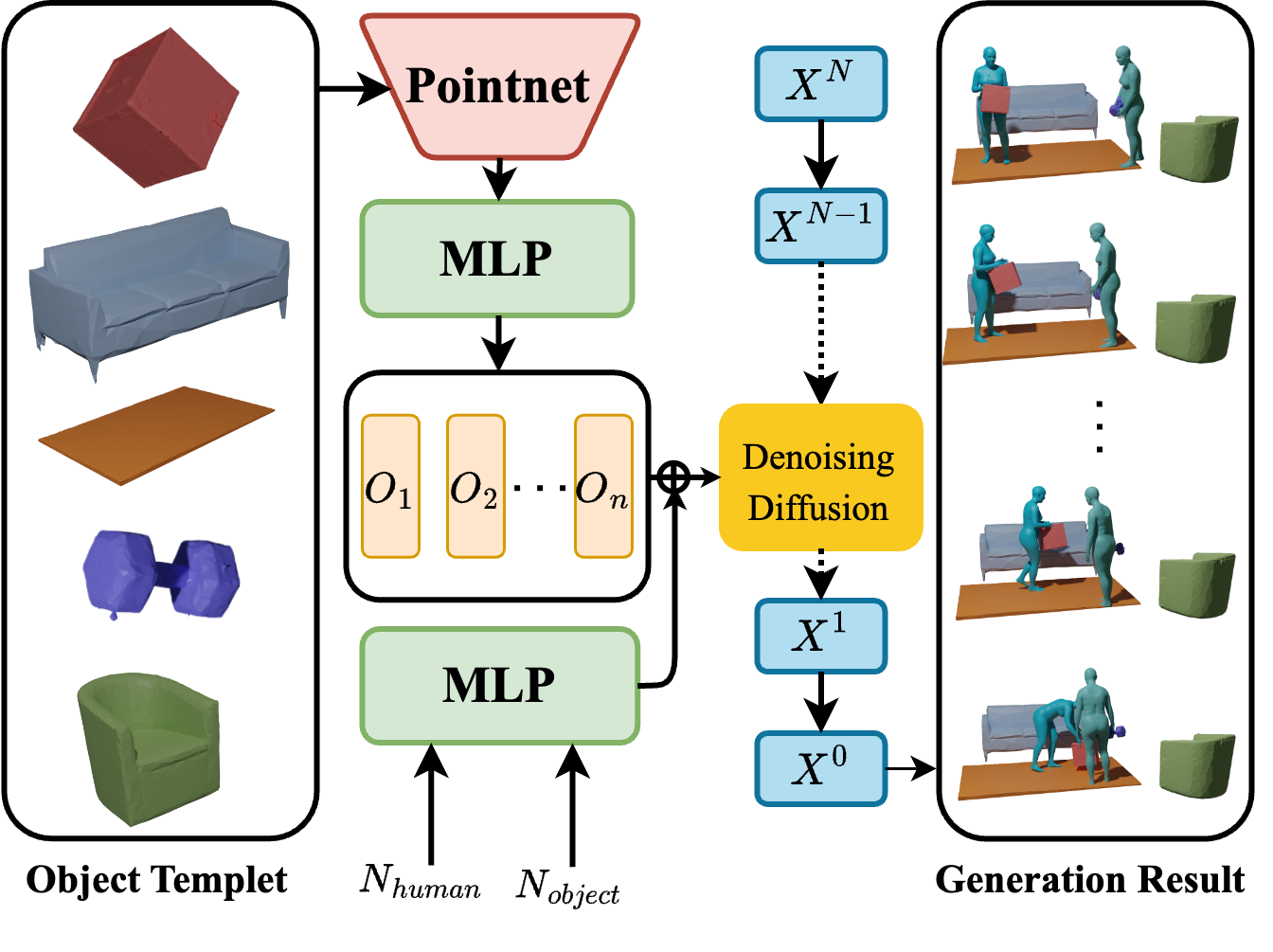}
    \captionof{figure}{\textbf{Multiple Interaction Generation Pipeline.} Given multiple object geometry, we employ Pointnet to extract the geometry features and feed them forward with the features of the preset number of humans and objects using an MLP. The resulting features are then fed into a conditional diffusion model to generate multiple human-object interactions.}
    \label{fig:diffusion}
\end{figure}

\begin{figure*}[t!]
  \centering
  \includegraphics[width=0.95\linewidth]{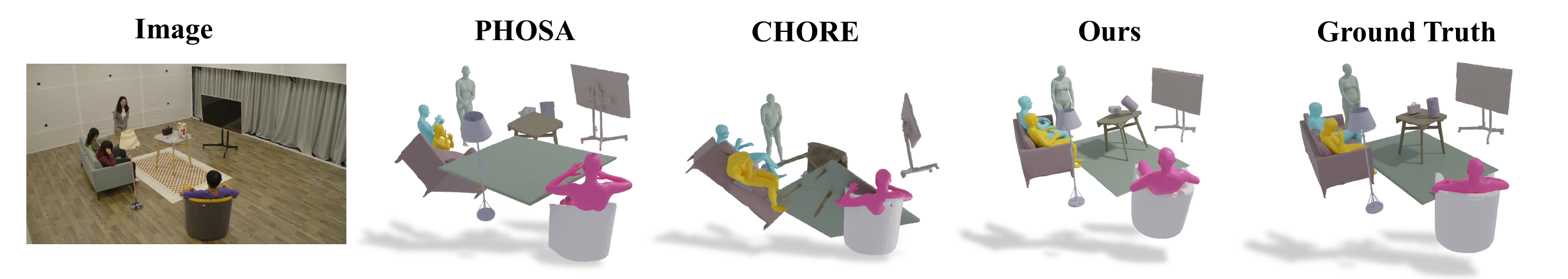}
    \captionof{figure}{Qualitative comparisons of monocular multiple interaction capture on HOI-M$^3$ dataset with two state-of-the-art monocular HOI capturing methods PHOSA~\cite{zhang2020phosa} and CHORE~\cite{xie2022chore}.}
  \label{fig:capture_result}
\end{figure*}


\noindent \textbf{Conditional Diffusion model.}
Referring to the typical implementations of denoising diffusion probabilistic models (DDPM) \cite{ho2020denoising} and Ego-Ego \cite{li2023ego}, the structure of the multiple interaction diffusion model is illustrated in Figure~\ref{fig:diffusion}. The high-level idea of the diffusion model is to design a forward diffusion process that adds Gaussian noises to the original data with a known variance schedule and learns a denoising model to gradually denoise $N$ steps given a sampled $x_{N}$ from a normal distribution to generate $x_0$.
Specifically, diffusion models comprise a forward diffusion process and a reverse diffusion process. The forward diffusion process gradually adds Gaussian noise to the original data $x_0$. It is formulated using a Markov chain of $N$ steps:


\begin{equation}
\begin{aligned}
q(x_{1:N} | x_0) := \prod_{n=1}^{N} q(x_n | x_{n-1}).
\end{aligned}
\end{equation}
Each step is decided by a variance schedule using $\beta_n$ and is defined as
\begin{equation}
\begin{aligned}
q(x_n | x_{n-1}) := \mathcal{N}(x_n ; \sqrt{1 - \beta_n} x_{n-1} , \beta_n \mathbf{I}),
\end{aligned}
\end{equation}
Learning the mean can be reparameterized as learning to predict the original data $x_0$. The training loss is defined as a reconstruction loss of $x_0$:
\begin{equation}
\begin{aligned}
\mathcal{L} = \mathbb{E}_{x_0, n} || \hat{x}_{\theta} (x_n, n) - x_0 ||_1.
\end{aligned}
\end{equation}
\begin{figure}[t]
    \centering
    \includegraphics[width=\linewidth]{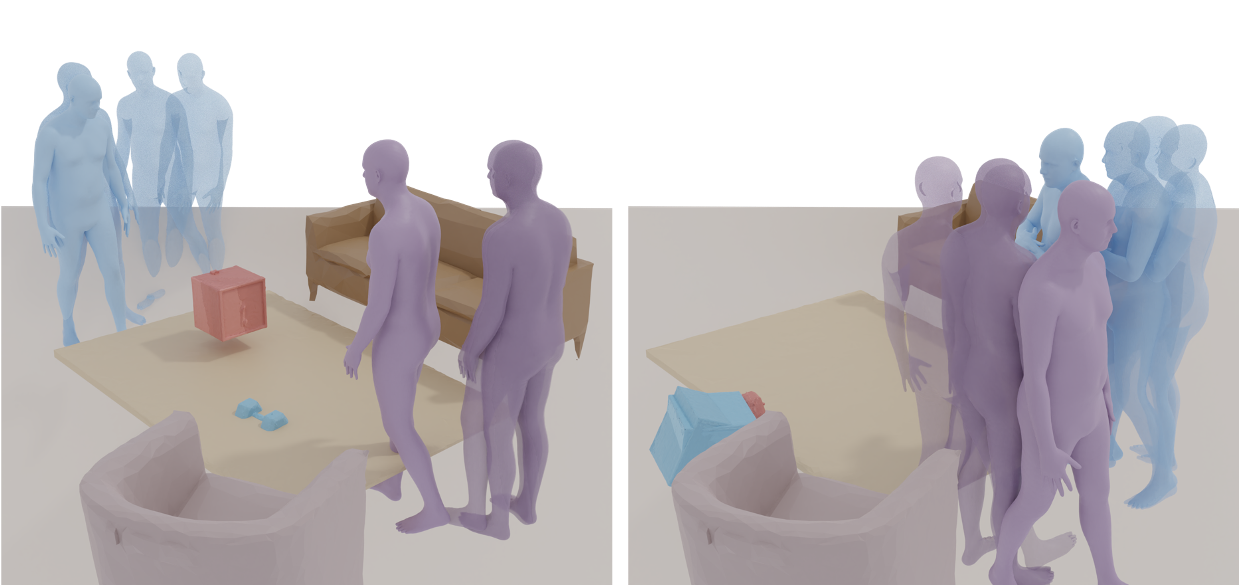}
    \captionof{figure}{Qualitative results of multiple interaction generation: We present the outcomes of two distinct sequences within a living room environment, each defined by specific object geometries and a predefined configuration of 2 persons and 5 objects.}
    \label{fig:diffusion_result}
\end{figure}
Here, we use the object geometry, the number of people and objects as conditions to generate the entire interaction. Thus, the number of people and objects is fed through an MLP as embedding to the network. The object geometry is extracted by Pointnet~\cite{qi2017pointnet} to obtain the global feature. Due to page limitations, we defer more details of the network structure to the Appendix.





\section{Experiments}

\begin{table*}[t!]
    \centering
    \small
    \setlength{\tabcolsep}{5pt}
    \begin{tabular}{lcccccccccc}
        \toprule
         & \multicolumn{4}{c}{ All } & \multicolumn{4}{c}{ Matched }\\
        \cmidrule(lr){2-5} \cmidrule(lr){6-9}
         & \multicolumn{2}{c}{Human} & \multicolumn{2}{c}{Object} & \multicolumn{2}{c}{Human} & \multicolumn{2}{c}{Object} \\
        \cmidrule(lr){2-3} \cmidrule(lr){4-5} \cmidrule(lr){6-7} \cmidrule(lr){8-9}
         Method & PCK$_{rel}$ $\uparrow$ & PCK$_{abs}$ $\uparrow$ & Chamfer$_{o}\downarrow$ & V2V$\downarrow$ & PCK$_{rel}$ $\uparrow$ & PCK$_{abs}$ $\uparrow$ & Chamfer$_{o}\downarrow$ & V2V$\downarrow$ \\
        \midrule
        PHOSA~\cite{zhang2020phosa} & 43.9 & - & 1454.3 & 691.4  & 48.8 & - & 1454.3 & 691.4 \\
        CHORE~\cite{xie2022chore} & 10.4 & - & 465.8 & 340.2  & 20.8 & -  & 465.8 & 340.2 \\
        Ours & \textbf{68.5} & \textbf{5.9} & \textbf{235.0} & \textbf{297.8} & \textbf{66.0} & \textbf{3.3} & \textbf{235.0} & \textbf{297.8} \\
        \bottomrule
    \end{tabular}
    \caption{Multiple HOI capture benchmark. ”Fit to input” represents the vanilla method that fits the object template to image and capture human with Frankmocap~\cite{rong2021frankmocap}.  The best results are in \textbf{bold}.}
    \label{tab:comparison}
\end{table*}

\begin{table}[t!]
    \centering
    \small
    \setlength{\tabcolsep}{4pt}
    \begin{tabular}{lccc} 
        \toprule
        & \multicolumn{2}{c}{Separated evaluation} & \multicolumn{1}{c}{Joint evaluation} \\
        \cmidrule(lr){2-3} \cmidrule(lr){4-4}
        Method & people & objects & Joint \\
        \midrule
        FID & 16.502 $\pm$ 0.044  & 10.609 $\pm$ 0.056   & 36.906 $\pm$ 0.087 \\ 
        Pene & 1.452$\%$  & 3.887$\%$ & 9.265$\%$  \\
        \bottomrule
    \end{tabular}
    \caption{Benchmark of multiple HOI generation on HOI-M$^{3}$. $\pm$ indicates the 95\% confidence interval.}
    \label{GenrationResult}
    \vspace{-1em} 
\end{table}

\subsection{Evaluation of the Multiple HOI Capturing}

We evaluate the proposed monocular multiple HOI capturing method on the HOI-M$^3$ dataset, and compare the evaluation result with two SOTA single HOI capturing methods~\cite{xie2022chore,zhang2020phosa}. We use the same input image size of 512×512 for all the methods to ensure a fair comparison.

\noindent \textbf{Datasets and Evaluation Metrics.} 
We train the Multiple HOI Capturing model using BEHAVE~\cite{bhatnagar2022behave}, InterCap~\cite{huang2022intercap}, and HOI-M$^3$, and perform evaluations on HOI-M$^3$. In this task, our goal is to estimate the pose of every human and object in camera-centric coordinates. To assess the accuracy of human poses, we employ the Percentage of Correct 3D Keypoints (PCK), which calculates the percentage of correct joints within 15cm of the ground truth joint location. For a more comprehensive evaluation of instant localization ability, we additionally employ 3DPCK$_\text{abs}$, which represents the 3DPCK without root alignment, assessing performance in absolute camera-centered coordinates~\cite{moon2019camera}. Regarding objects, we use chamfer distance and mean vertex to vertex(v2v) to assess the accuracy of the object's results. It's important to note that by 'match,' we specifically mean that we consider accuracy only for matched ground truths.

\noindent \textbf{Monocular Multiple HOI Capturing Benchmark} 
We compare our evaluation results with two state-of-the-art single HOI capturing methods~\cite{xie2022chore, zhang2020phosa}. While these methods are designed for single HOI cases, we compute the Intersection over Union (IOU) for each bounding box. Then we select the human-object pair with the best IOU to obtain their results. From Tab.~\ref{tab:comparison}, our proposed multiple HOI capturing significantly surpasses existing methods. We observe that the weak-projection camera model used in current single HOI methods leads to inaccuracies in root depth. Consequently, we are unable to calculate the PCK${\text{abs}}$ for these two methods. Nevertheless, our method also demonstrates superiority in PCK${\text{rel}}$, highlighting its local pose estimation capabilities. Regarding objects, our method exhibits lower chamfer distance compared to PHOSA and CHORE. It is noteworthy that the aforementioned methods require the presetting of the number of objects, resulting in identical performance for both match and all predictions. We also show the qualitative comparison in Figure~\ref{fig:capture_result}, where it is clear that, despite our method showing superior human and object quantitative results, capturing vivid motions of multiple human-object interactions remains a challenging direction.

\subsection{Evaluation of the Multiple HOI Generation}

\noindent \textbf{Evaluation Metrics.} 
We introduce two metrics, FID and Pene, to evaluate this novel task. 1) FID is a metric used to assess the quality of the generated image by comparing the differences in the distribution of feature vectors extracted from the generated and real images using Inception v3 models. The results demonstrate the remarkable performance of our generation output. 2) Pene measures the average percentage of object vertices with non-negative human signed distance function values.

\noindent \textbf{Multiple HOI Generation Benchmark} 
We evaluate our model based on 20 sampling. The result shows in Tab.~\ref{GenrationResult}. For a more intuitive comparison, we provide the visual results of the generated motion in Figure~\ref{fig:diffusion_result}, where we can clearly see that the model trained on HOI-M$^3$ can synthesize semantically corresponding motions given object inputs and specify the number of people and object. These results prove the significant advantages of our dataset in generating such diverse social interaction.

\subsection{Limitations}
While HOI-M$^3$ is the first to provide possibilities for exploring varied relationships between interacting subjects, equipped with capturing label of multiple persons and multiple objects within an environment, we also want to highlight some potential limitations of this direction. Firstly, due to hardware cost constraints, HOI-M$^3$ is currently limited to indoor settings, and extending the current setup to outdoor environments, particularly in the wild, poses non-trivial challenges. Secondly, building such a dataset involves significant human resources; thus, HOI-M$^3$ only covers 5 common scenes. Moreover, our dataset was collected under fixed illumination conditions with few background variations, limiting its generalization ability to other environments.






\section{Conclusion}

We have introduced HOI-M$^3$, a pioneering dataset designed for capturing interactions involving multiple humans and objects within a contextual environment. Key features of our HOI-M$^3$ dataset include: 1) Multiple Humans and Objects, 2) high quality, and 3) large size with rich modalities. Leveraging our dataset, we meticulously devised two robust baseline methods for downstream tasks: monocular capture of multiple HOI and generation of multiple HOI. We conduct comprehensive evaluations of our dataset and companion baseline methods, presenting preliminary results to indicate that capturing or generating vivid motions of multiple human-object interactions remains a challenging research direction.
We expect that this research will boost the advancement in the context of multiple HOI.



\noindent\textbf{Acknowledgement}
This work was supported by the Shanghai Local College Capacity Building Program (23010503100,22010502800), Shanghai Sailing Program (21YF1429400, 22YF1428800),  NSFC programs (61976138, 61977047), the National Key Research and Development Program (2018YFB2100500), STCSM (2015F0203-000-06), SHMEC (2019-01-07-00-01-E00003), Shanghai Advanced Research Institute, Chinese Academy of Sciences, Shanghai Engineering Research Center of Intelligent Vision and Imaging and Shanghai Frontiers Science Center of Humancentered Artificial Intelligence (ShangHAI).
\clearpage
\renewcommand\thesection{\Alph{section}}
\setcounter{section}{0}

\section{More Details of HOI-M$^3$ Dataset}
\label{sec:Dataset}

In this section, we provide more details about HOI-M$^3$ dataset, including statistic analyses, data preprocess and hardware setup.

\subsection{Dataset Statistic}
\begin{figure*}[t!]
  \centering
  \includegraphics[width=1.0\linewidth]{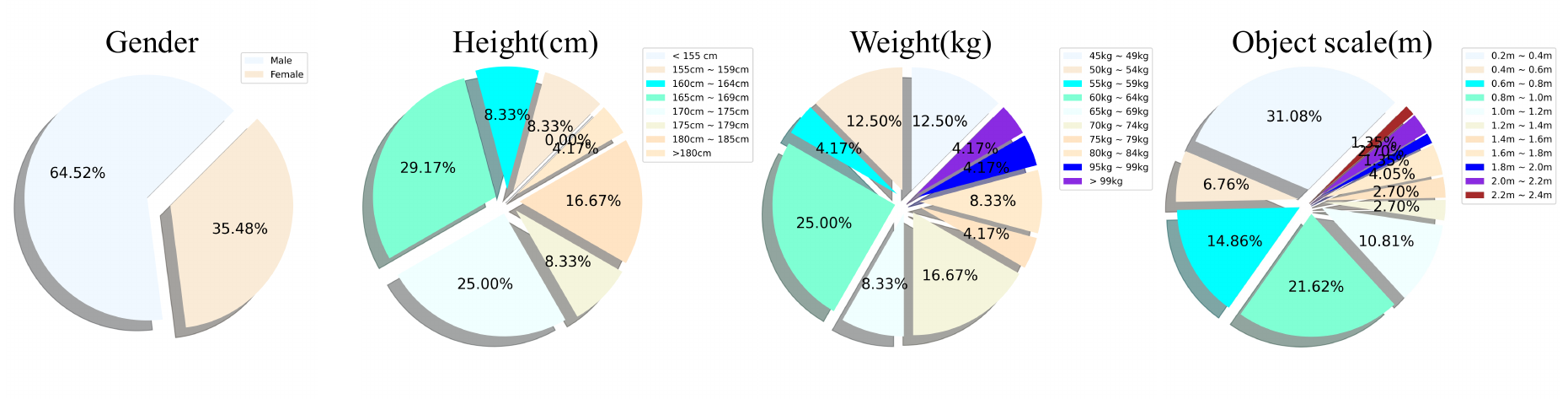}
 \caption{Statistics of HOI-M$^3$ humans and objects.}
  \label{fig:Statistic}
\end{figure*}

HOI-M$^3$ provides a large volume of long human object interactions(HOI) (more than 10k frames HOI per sequences), which will be beneficial for long-term motion and HOI generation.  To assess the dataset's diversity, we provide key statistics, including gender, height, weight, and object scale, illustrated in Figure~\ref{fig:Statistic}. The results demonstrate the dataset's diversity in human body shapes and object scales.

\subsection{Data Preprocess}

For accurate object tracking, separating the target object from the background in a video sequence serves as a crucial cue for optimization. However, tracking an arbitrary object in diverse scenes is a non-trivial task. Following previous work, Track-Anything~\cite{yang2023track}, we employ the Segment Anything Model (SAM)\cite{kirillov2023segment} to annotate the initial frame of each camera view. Subsequently, we utilize XMem\cite{cheng2022xmem} for video object tracking (VOS) on the subsequent frames.

\subsection{Hardware Setup}

\begin{figure*}[t!]
  \centering
  \includegraphics[width=1.0\linewidth]{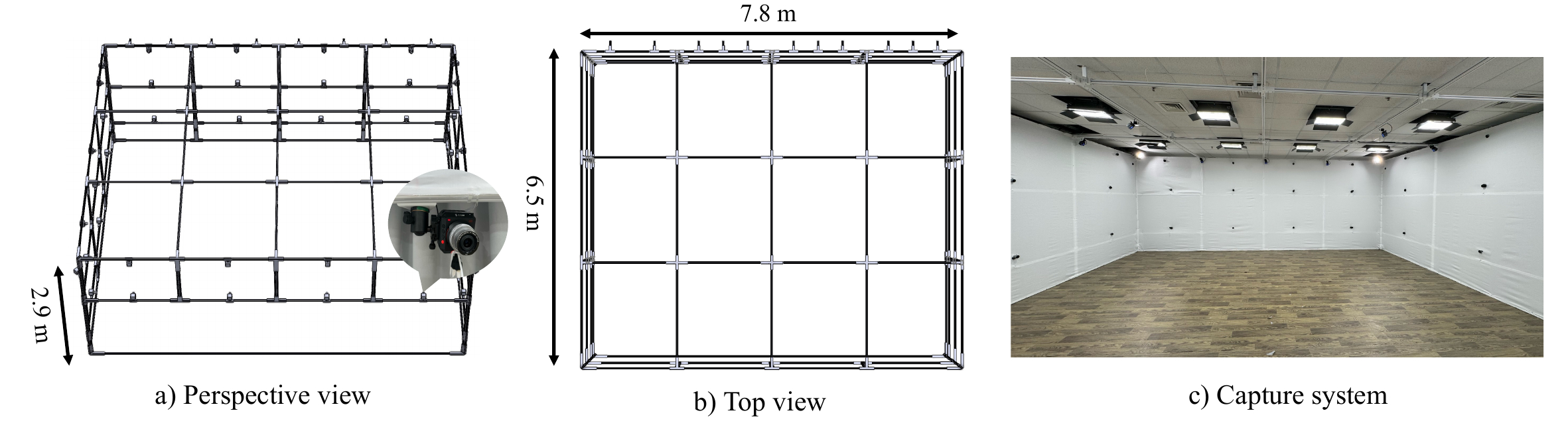}
 \caption{Hardware setup.}
  \label{fig:framework}
\end{figure*}

\begin{figure*}[t!]
  \centering
  \includegraphics[width=1.0\linewidth]{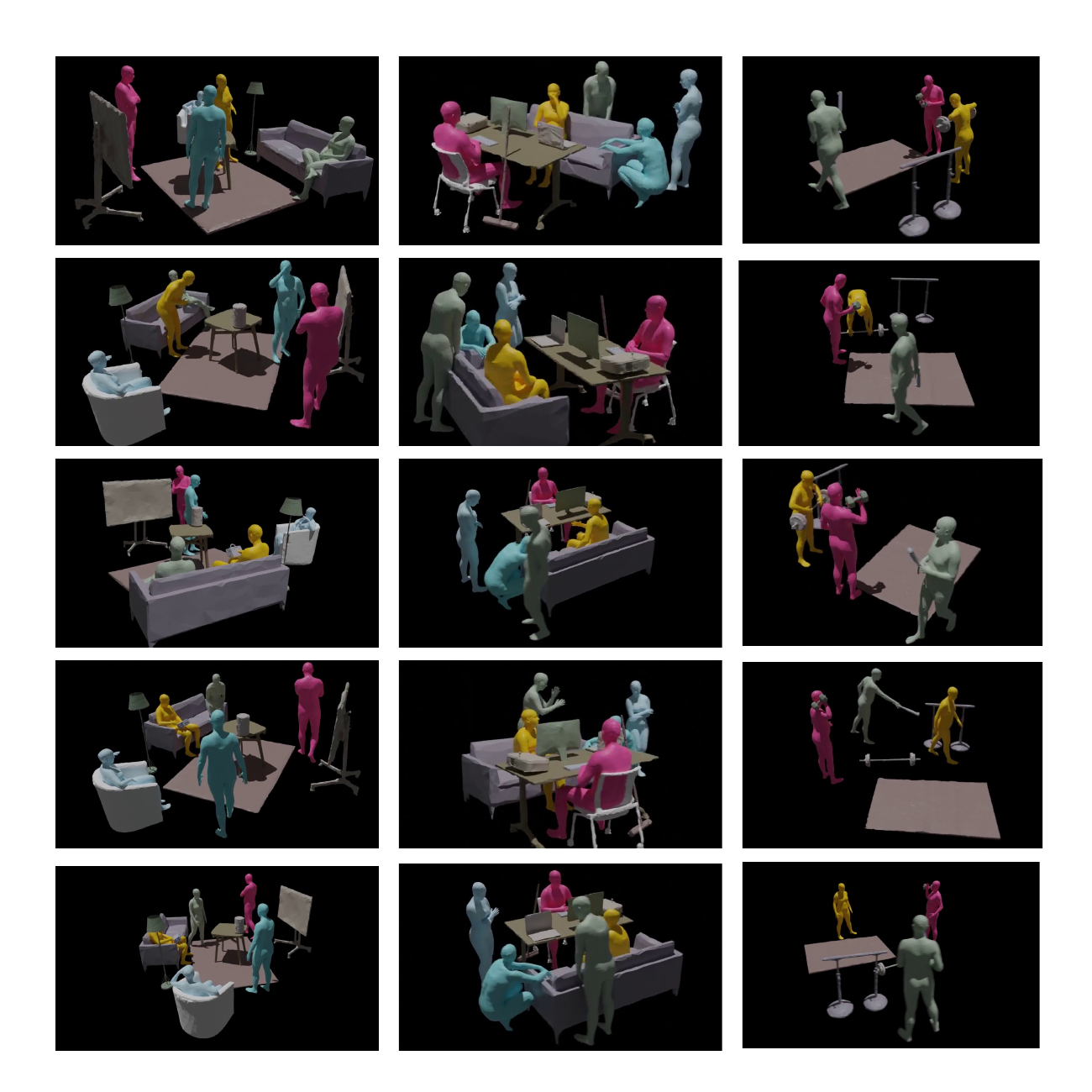}
 \caption{More quality results.}
  \label{fig:morequality}
\end{figure*}

Accurately capturing the motions of multiple humans and objects remains a challenging task, particularly in the presence of severe occlusions, a common occurrence in daily interactions within contextual environments. To address this challenge and capture realistic interaction sequences, we designed a custom room-like dome with a square-shaped multi-layer framework to house the RGB sensors. The system stands at a height of 2.9 m and has a side length of 7.8 m for its octagonal cross-section, as illustrated in Figure~\ref{fig:framework}. To better align our capture setup with everyday scenarios, we opted for white backdrops instead of green ones to conceal the cable. We also provide more quality results sampled from HOI-M$^3$ dataset as shown in Figure~\ref{fig:morequality}.

\section{How HOI-M$^3$ Contributes to the Community?}


The HOI-M$^3$ dataset comprises various scenes depicting human-object interactions, accompanied by per-frame multiple human and object tracking. We believe our dataset addresses a significant gap in the literature on multiple human-object interactions. At the meanwhile, we anticipate that the dataset will serve as a valuable resource for various research directions. We propose the following challenges based on the HOI-M$^3$ dataset:

\myparagraph{Multiple Person Pose and Shape Estimation.}
HOI-M$^3$ offers parametric model labels encompassing shape information and 3D skeletal positions. This provides a robust benchmark for multi-person scenarios, particularly in daily situations where individuals are frequently occluded by surrounding objects. We believe that HOI-M$^3$ serves as a reflective measure of each method's performance in such challenging scenarios.

\myparagraph{Multiple HOI Capture.}
In recent years, significant advancements have been made in data-driven human motion capture, even for single HOI capture. However, there has been limited progress in monocular multiple HOI capture. The HOI-M$^3$ dataset addresses this gap by providing the largest and most accurate capturing labels paired with natural RGB images, enabling robust HOI supervision. Consequently, our dataset is well-suited for data-driven approaches in both monocular and multi-view settings, leveraging the precision of our ground truth annotations.

\myparagraph{Multiple Human Motion Generation.}
We have witnessed remarkable advancements in diffusion techniques for generating lifelike human motions, progressing from single human motion~\cite{tevet2022human, dabral2022mofusion, karunratanakul2023gmd, zhang2020generating, zhang2020place, zhang2021we} to the recent exploration of two-human interactions~\cite{liang2023intergen}. Leveraging the extensive dataset of long-duration multi-human motions in HOI-M$^3$, we can offer accurate labels for multi-human interactions to facilitate this evolving task.

\myparagraph{Multiple Interaction Generation.}
HOI-M$^3$ provides an extensive collection of diverse interaction sequences with synchronized ground truth capture. Motivated by the recent significant progress in Motion Generation (MoGen) tasks, we have demonstrated how our dataset contributes to this field in the main paper, particularly in the context of a novel task: Multiple Interaction Generation.

\section{More Details of Monocular Multiple HOI Capture}
\label{sec:Capture}

\subsection{Network Architecture}
For a fair comparison, we do not choose large size of backbone; instead, we employ ResNet-34~\cite{he2016deep}, pre-trained on the ImageNet dataset~\cite{deng2009imagenet}, as the default backbone. All input images were padded to the standardized size of $512 \times 512$. Each prediction head attached to the backbone comprises a $3 \times 3 \times 256$ convolutional layer, BatchNorm, ReLU, and another $1 \times 1 \times c_0$ convolutional layer, where $c_0$ represents the output size.


\subsection{Loss Function}

To supervise the network, we have developed individual loss functions for different maps. The network is supervised by the weighted sum of the body pose loss $L_{\text{theta}}$, the body shape loss $L_{\text{beta}}$, the object pose loss $L_{\text{object}}$, the 3D keypoints loss $L_{\text{3D}}$, the 2D keypoints loss $L_{\text{2D}}$, the center keypoint heatmap $L_{\text{hm}}$, and the depth loss of humans and objects $L_{\text{depth}}$.

\myparagraph{Human Object Center Loss.}
We employ a heatmap representing the 2D human body center and object center in the image, which is represented as a Gaussian distribution in the human-object position. The center keypoint heatmap $L_{\text{hm}}$ is derived as follows:
\begin{equation}
\begin{aligned}
L_{\text{hm}} & = \| C_m^{\text{pred}} - C_m^{\text{gt}}  \|_2,
\end{aligned}
\end{equation}
where $C_m^{\text{pred}} \in \mathbb{R}^{128 \times 128}$ is the predicted center heatmap, and $C_m^{\text{gt}}  \in \mathbb{R}^{128 \times 128}$ is the ground truth of $C_m^{\text{pred}}$.


\myparagraph{Human Parameter Loss.}
Through the parameter sampling process, we enforce the human parameter loss $L_{\text{theta}}$ and $L_{\text{beta}}$ to match each ground truth body with a predicted parameter result for supervision. The body pose loss $L_{\text{theta}}$ and the body shape loss $L_{\text{beta}}$ are derived as follows:
\begin{equation}
\begin{aligned}
L_{\text{theta}} & = \| \theta^{\text{pred}} - \theta^{\text{gt}}  \|_1, \\
L_{\text{beta}} & = \| \beta^{\text{pred}} - \beta^{\text{gt}}  \|_1,
\end{aligned}
\end{equation}
where $\theta^{\text{gt}}  \in \mathbb{R}^{24\times 3}$ and $\beta^{\text{gt}}  \in \mathbb{R}^{10}$ denote the ground truth of the model's parameters. $\theta^{\text{pred}}  \in \mathbb{R}^{24\times 3}$ and $\beta^{\text{pred}} \in \mathbb{R}^{10}$ denote the predicted parameter results sampled from each center position of the human. Here we use the $\ell_1$ norm, following previous work~\cite{zhang2021body,sun2021monocular}.

\myparagraph{Object Pose Loss.}
Similar to Human Parameter, we sample the object's 6D pose from each object center with a predicted parameter result for supervision. The object pose loss $L_{\text{object}}$ is derived as follows:
\begin{equation}
\begin{aligned}
L_{\text{object}} & = \| R^{\text{pred}} - R^{\text{gt}}  \|_1,
\end{aligned}
\end{equation}
where $R^{\text{pred}}  \in \mathbb{R}^{3\times2}$ denotes predicted object rotation, and $R^{\text{gt}}\in \mathbb{R}^{3\times2}$ denotes the ground truth of the rotation.

\myparagraph{Depth Loss.}
Besides the local representation of humans and objects, another key component is depth. Here we impose each subject's depth as follows:
\begin{equation}
\begin{aligned}
L_{\text{object}} & = \| Z_{\text{center}}^{\text{pred}} - Z_{\text{center}}^{\text{gt}}  \|_1,
\end{aligned}
\end{equation}
where $Z_{\text{center}}^{\text{pred}}\in \mathbb{R}$ denotes the predicted depth of humans or objects, and $Z_{\text{center}}^{\text{gt}} \in \mathbb{R}$ denotes the ground truth of the depth.

\myparagraph{Additional Loss.}
In addition to imposing supervision on each regression target, we also utilize some intermediate supervised signals for training, such as 2D keypoints and 3D keypoints of humans:
\begin{equation}
\begin{aligned}
L_{\text{2D}} & = \| P_{\text{2D}}^{\text{pred}} - P_{\text{2D}}^{\text{pred}}\|_1,\\
L_{\text{3D}} & = \| P_{\text{3D}}^{\text{pred}} - P_{\text{3D}}^{\text{pred}}\|_1,
\end{aligned}
\end{equation}
where $ P_{\text{2D}}^{\text{pred}} \in \mathbb{R}^{24\times 2}$ and $ P_{\text{3D}}^{\text{pred}}\in \mathbb{R}^{24\times 3}$ denote predicted 2D and 3D keypoints, and $ P_{\text{2D}}^{\text{gt}} \in \mathbb{R}^{24\times 2}$ and $ P_{\text{3D}}^{\text{gt}} \in \mathbb{R}^{24\times 3}$ denote the ground truth of 2D and 3D keypoints.

\begin{figure}[t]
    \centering
    \includegraphics[width=\linewidth]{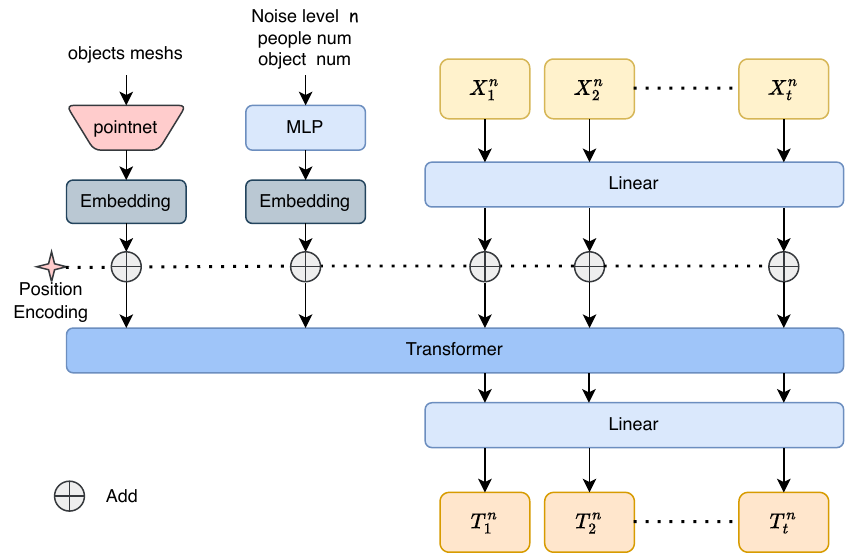}
    \captionof{figure}{Model architecture of denoising network.}
    \label{fig:diffusion_moredetail}
\end{figure}


\section{More Details of Multiple Interaction Generation}
\label{sec:Generation}

Our diffusion models encompass both a forward diffusion process and a reverse diffusion process. The forward diffusion process progressively introduces Gaussian noise to the original data $x_0$. In this case, we employed a transformer model architecture as our denoising network, comprising four self-attention blocks. Each self-attention block consists of a multi-head attention layer followed by a position-wise feed-forward layer. Illustrated in Figure~\ref{fig:diffusion_moredetail}, our denoising network incorporates several feature embeddings. Specifically, it includes embeddings from object meshes and condition signals of noise levels $n$, human numbers, and object numbers, which are then concatenated together as input to our transformer model.

\section{Experiment}

\subsection{Ablation Study}

To comprehensively evaluate the components of inertial-aided multi-object tracking, we perform an additional qualitative analysis of various constraint terms. It is important to note that we lack ground truth specific to tracking, so our evaluations are qualitative in nature. In figure~\ref{fig:ablation}, we present the quality results obtained by ablating different components. Specifically, "w/o collision," "w/o IMU Init," and "w/o offscreen loss" denote the results obtained without using the collision constraint term, without employing the IMU as initialization, and without utilizing the offscreen term $E_{\text{offscreen}}$, respectively. The results demonstrate that the offscreen term $E_{\text{offscreen}}$ effectively prevents degenerate results. Furthermore, without IMU initialization, recovering the object's rotation from the human-object mask becomes challenging, and our collision loss ensures realistic interactions between humans and objects.

\subsection{More Benchmarks}
\noindent\textbf{Monocular 3D Human Pose and Shape Estimation} 
In addition to the two benchmarks for novel data-driven tasks and their corresponding strong baselines presented in the main paper, we also introduce additional benchmarks for a prevalent vision task: monocular 3D human pose and shape estimation. To ensure a fair comparison with existing works, we conduct several experiments on our datasets. For evaluation metrics, we utilize mean per joint position error (MPJPE), procrustes aligned mean per joint position error (PA-MPJPE), the percentage of correct keypoints (3DPCK), and area under curve (3D-AUC) to assess the performance of 3D pose due to their common usage. Additionally, we employ per vertex error (PVE) to evaluate body mesh estimation ability. Furthermore, we report the percentage of correct keypoints after procrustes alignment (PA-3DPCK) and area under curve after procrustes alignment (PA-3DAUC) on our dataset. We believe that our dataset currently stands as the most comprehensive benchmark in terms of evaluation metrics. The main results are presented in Table~\ref{tab:comparison}, indicating that conducting tests in scenarios involving multiple persons within multiple object occlusions poses a significant challenge compared to results obtained from other datasets.


\begin{figure*}[t!]
  \centering
  \includegraphics[width=1.0\linewidth]{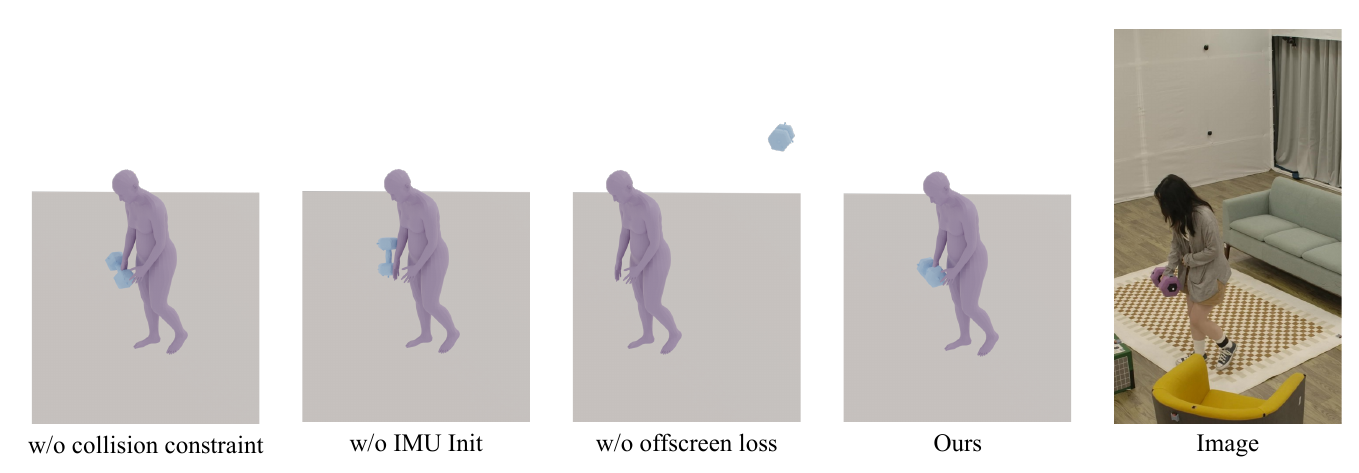}
 \caption{Qualitative evaluation.}
  \label{fig:ablation}
\end{figure*}

\begin{table*}[t!]
    \centering
    \small
    \setlength{\tabcolsep}{10pt}
    \begin{tabular}{lcccccccccc}
        \toprule
         Method & MPJPE$\downarrow$ & PA-MPJPE$\downarrow$ & 3DPCK$\uparrow$ & PA-3DPCK $\uparrow$ & 3DAUC$\uparrow$ & PA-3DAUC$\uparrow$ & PVE$\downarrow$ \\
        \midrule
        HMR~\cite{kanazawa2018end} & 324.60 & 187.69 & 13.64 & 50.68  & 3.57 & 16.71 & 404.49 \\
        SPIN~\cite{kolotouros2019learning} & \textbf{309.81} & 160.56 & 16.97 & 53.74  & 5.11 & 23.33 & 357.00 \\
        HybrIK~\cite{li2021hybrik} & 326.86 & \textbf{127.74} & \textbf{18.95} & \textbf{68.79}  & \textbf{6.74} & \textbf{29.96}  & \textbf{335.55} \\
        PARE~\cite{Kocabas_PARE_2021} & 325.64 & 188.65 & 9.63 & \textbf{46.50} & 1.77 & 14.49 & 403.25 \\
        BalancedMSE~\cite{ren2021bmse} & 331.93 & 152.40 & 13.85 & 56.35 & 4.02 & 26.06 & 346.16 \\
        CLIFF~\cite{li2022cliff} & 332.47 & 161.09 & 14.31 & 58.39 & 4.25 & 23.54 & 413.70 \\
        \bottomrule
    \end{tabular}
    \caption{Monocular 3D human pose and shape estimation benchmark. The best results are in \textbf{bold}.}
    \label{tab:comparison}
\end{table*}

\begin{figure*}[t!]
  \centering
  \includegraphics[width=1.0\linewidth]{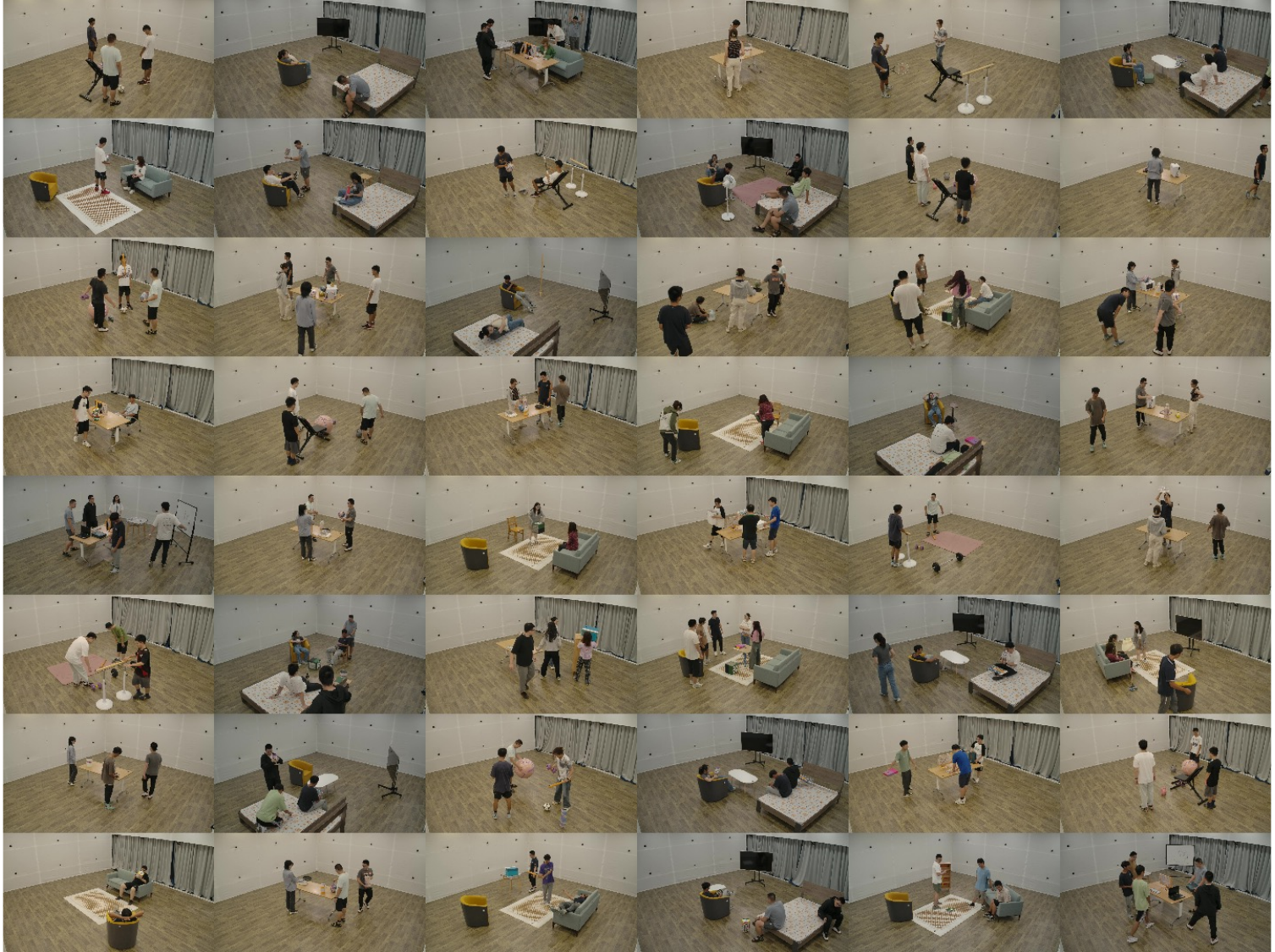}
 \caption{Data examples were captured by our system.}
  \label{fig:ablation}
\end{figure*}

{
    \small
    \bibliographystyle{ieeenat_fullname}
    \bibliography{main}
}


\end{document}